\documentclass{article}

\usepackage{amsmath,amsfonts,bm}

\def\eqref#1{equation~\ref{#1}}

\def\1{\bm{1}}

\DeclareMathAlphabet{\mathsfit}{\encodingdefault}{\sfdefault}{m}{sl}
\SetMathAlphabet{\mathsfit}{bold}{\encodingdefault}{\sfdefault}{bx}{n}

\usepackage[nonatbib,preprint]{neurips_data_2024}

\usepackage[utf8]{inputenc} 
\usepackage[T1]{fontenc}    
\usepackage{url}            
\usepackage{booktabs}       
\usepackage{amsfonts}       
\usepackage{nicefrac}       
\usepackage{microtype}      
\usepackage[dvipsnames, svgnames, x11names]{xcolor}
\usepackage{booktabs}
\usepackage{multirow}
\usepackage{makecell}
\usepackage{cellspace}
\usepackage{graphicx}
\usepackage{colortbl}
\usepackage[shortlabels]{enumitem}
\usepackage{bbding}
\usepackage{caption}
\usepackage{subcaption}
\usepackage[symbol]{footmisc}
\usepackage[colorlinks=true,urlcolor=blue,linkcolor=black,citecolor=black]{hyperref}
\usepackage{cleveref}
\usepackage{wrapfig}
\usepackage{pifont}
\definecolor{'wit'}{HTML}{FBFBFB}
\definecolor{'gry'}{HTML}{EEEEEE}
\definecolor{mypink}{rgb}{.99,.91,.95}
\definecolor{'deep1'}{HTML}{C5E6F8} 
\definecolor{'shallow1'}{HTML}{E4F3FC} 
\definecolor{'deep2'}{HTML}{E5F5B7} 
\definecolor{'shallow2'}{HTML}{F3FADF} 

\definecolor{'deep3'}{HTML}{FFE5C6} 
\definecolor{'shallow3'}{HTML}{FFF2E3} 
\definecolor{'deep4'}{HTML}{FFD3CF} 
\definecolor{'shallow4'}{HTML}{FFEAE8}
\definecolor{'deep5'}{HTML}{D2D0F3} 
\definecolor{'shallow5'}{HTML}{E8E7F9} 
\crefname{section}{§}{§§}
\Crefname{section}{§}{§§}
\crefname{figure}{Figure}{Figure}
\Crefname{figure}{Figure}{Figure}
\crefname{table}{Table}{Table}
\Crefname{table}{Table}{Table}
\Crefname{appendix}{Appendix}{Appendix}
\crefname{appendix}{Appendix}{Appendix}
\crefname{equation}{Eq.}{Eq.}

\title{
\includegraphics[scale=0.025]{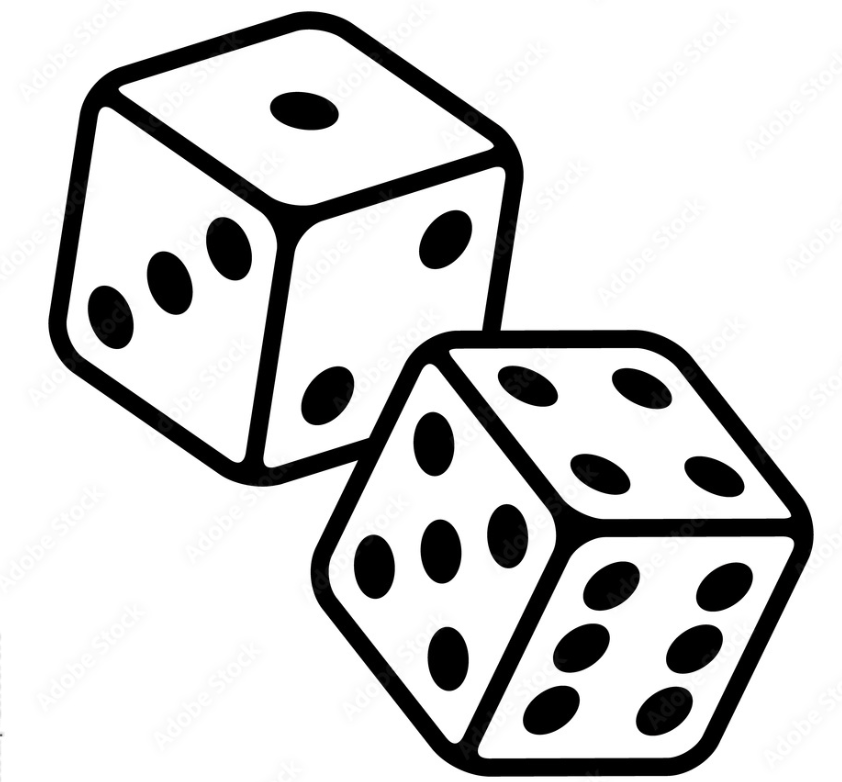}
DICE: Detecting In-distribution Contamination  \\ in LLM's Fine-tuning Phase for Math Reasoning
}

\author{%
  \centerline{Shangqing Tu$^{1}$\footnotemark[1], ~Kejian Zhu$^{2}$\footnotemark[1], ~Yushi Bai$^1$, ~Zijun Yao$^1$, ~Lei Hou$^1$, ~Juanzi Li$^1$\footnotemark[2]} \\
\centerline{$^1$Tsinghua University \quad $^2$Beihang University}\\
\centerline{\texttt{\{tsq22,bys22\}@mails.tsinghua.edu.cn}, \texttt{\{houlei, lijuanzi\}@tsinghua.edu.cn}}
}

\begin{document}

\maketitle
\renewcommand{\thefootnote}{\fnsymbol{footnote}}
\footnotetext[1]{Equal Contribution.}
\footnotetext[2]{Corresponding author.}
\renewcommand*{\thefootnote}{\arabic{footnote}}

\begin{abstract}

The advancement of large language models (LLMs) relies on evaluation using public benchmarks, but data contamination can lead to overestimated performance.
Previous researches focus on detecting contamination by determining whether the model has seen the exact same data during training. Besides, prior work~\cite{dekoninck2024evading} has already shown that even training on data similar to benchmark data inflates performance, namely \emph{In-distribution contamination}. In this work, we argue that in-distribution contamination can lead to the performance drop on OOD benchmarks.
To effectively detect in-distribution contamination, we propose DICE, a novel method that leverages the internal states of LLMs to locate-then-detect the contamination.
DICE first identifies the most sensitive layer to contamination, then trains a classifier based on the internal states of that layer.
Experiments reveal DICE's high accuracy in detecting in-distribution contamination across various LLMs and math reasoning datasets.
We also show the generalization capability of the trained DICE detector, which is able to detect contamination across multiple benchmarks with similar distributions.
Additionally, we find that DICE's predictions correlate with the performance of LLMs fine-tuned by either us or other organizations, achieving a coefficient of determination ($R^2$) between 0.61 and 0.75. \\
{\color{violet} \textbf{Code \& Data}}: \url{https://github.com/THU-KEG/DICE}.

\end{abstract}

\section{Introduction}
\label{sec:introduction} 


Current development of large language models (LLMs) and their related techniques heavily relies on public benchmarks to ensure that progress is made in the right direction. For instance, the field primarily uses GSM8k~\cite{gsm8k} and MATH~\cite{hendrycks2021measuring} to evaluate LLMs' math reasoning abilities.
However, there is growing concern that some of the impressive performance on these benchmarks may be attributed to data contamination, where the training data contains original data from the benchmark and is memorized by the model~\cite{xu2024benchmarking}, which we refer to as \emph{Exact Contamination}.

Recently, EAL~\cite{dekoninck2024evading} has already shown that training on data that bears similarity to the benchmark data can lead to severe performance overestimation, namely \emph{In-distribution contamination}. Since pre-training data is massive and hard to distinguish based on their distributions, we narrow our scope to the supervised fine-tuning (SFT) phase.
We aim to answer the following research questions:
(1) Does in-distribution contamination contribute to a model's overall math reasoning ability?
(2) If not, how can we detect it to prevent overestimating the model's capabilities due to contamination?

To investigate whether in-distribution contamination can really improve LLM's math reasoning ability, we design an OOD test for a set of fine-tuned LLMs simulating different levels of in-distribution contamination on GSM8K following prior work's experiment setup~\cite{dekoninck2024evading}.
In section \ref{sec:benchmark_alignment_tax}, we show that even if the contamination data in the training data accounts for only a small portion, the models' gain on in-distribution (ID) benchmarks is significantly greater than the gain on OOD benchmarks. The result suggests that ID data does not truly enhance the model's overall math reasoning ability, and the performance on in-distribution benchmarks no longer reflect the model's real capabilities.
This motivates the need for in-distribution contamination detection: Given a fine-tuned LLM on a piece of test data, determine if the model has seen its in-distribution data during fine-tuning.

\begin{figure}[t]
    \centering
    \includegraphics[width=\linewidth]{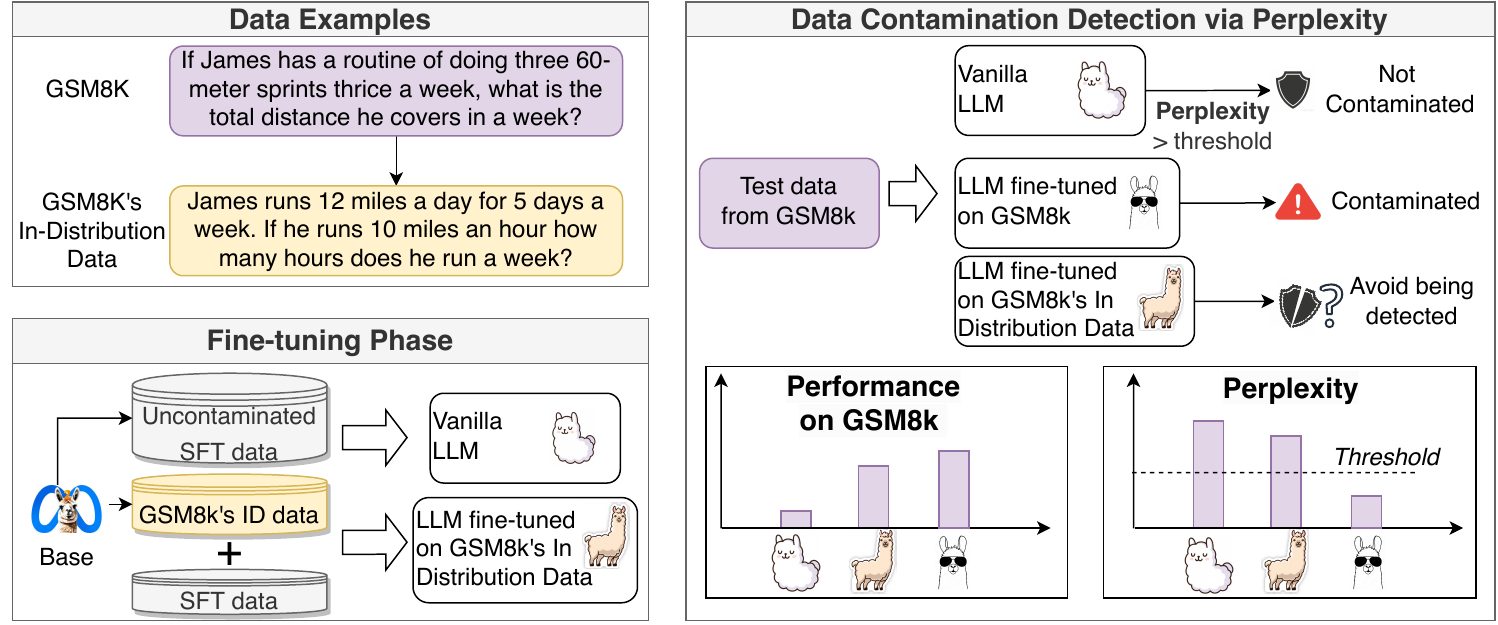}
    \caption{\textbf{Traditional contamination detection methods cannot handle in-distribution contamination}. Vanilla LLM refers to the LLM fine-tuned with uncontaminated data.}
    \label{fig:kola_framework}
\vspace{-1em}
\end{figure}

As shown in Figure~\ref{fig:kola_framework}, previous methods~\cite{hendrycks2020measuring,ji2023survey} detect contamination by measuring the model's memorization level of the test data, \textit{e.g.,} by evaluating the perplexity on the test data to detect whether the model has seen the original test data. 
However, unlike exact contamination, the in-distribution contaminated training data, such as being rewritten from the test data, only bears similarity with the test data at a semantic level.
This results in the failure of all previous contamination detection methods, as illustrated in the lower right of Figure~\ref{fig:kola_framework}.


\begin{figure}[t]
  \centering
  \includegraphics[width=\linewidth]{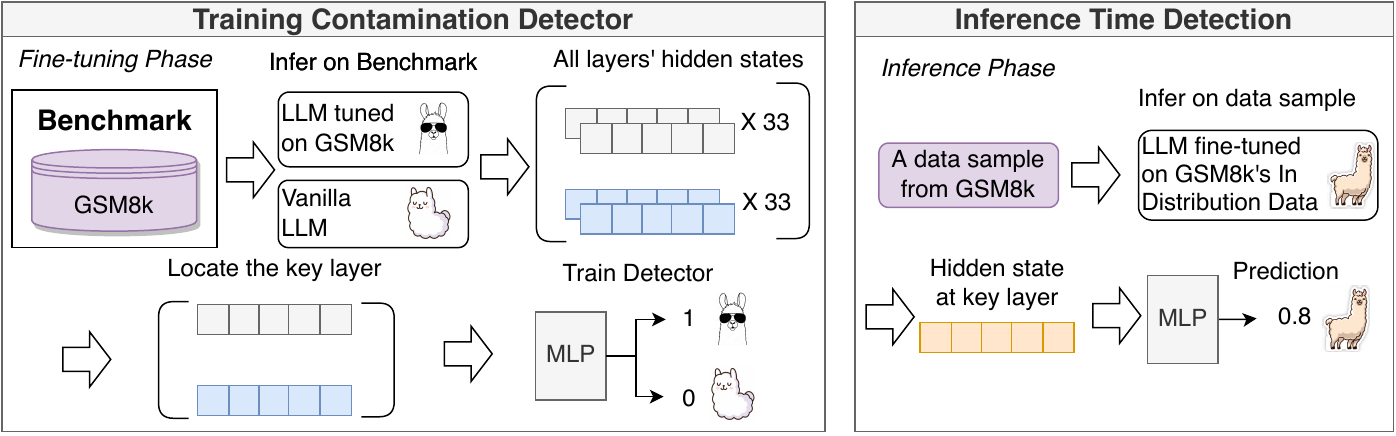}
  \caption{\textbf{Overview of DICE}. Locate-then-Detect LLM's in-distribution data contamination.}
  \label{fig:our_approach}
\end{figure}

To \underline{d}etect \underline{i}n-distribution \underline{c}ontamination in the fin\underline{e}-tuning phase of LLMs, we propose a novel locate-then-detect method called DICE, which
locates evidence in the internal states of LLMs for detecting their contamination.
As shown in Figure \ref{fig:our_approach}, we first fine-tune the base LLM on datasets with different levels of contamination to get different reference models. Next, we locate the \textit{contamination layer}, \textit{i.e.,} the most sensitive layer to the level of contamination, 
by identifying the layer with the maximum state distance between the contaminated and uncontaminated models.
Finally, we train an MLP as the classifier to quantify the contamination level based on the contamination layer's states.

Extensive experiments on different LLMs and datasets show that DICE accurately detects in-distribution contamination in LLMs. 
Moreover, we find that the DICE detector trained on GSM8K can generalize well to detecting contamination on other in-distribution datasets such as GSM-hard or paraphrased GSM8K. 
We also find that DICE's predictions highly correlate with the performance of LLMs on in-distribution and out-of-distribution tasks, even though our DICE detector only use input sequences' hidden states. Therefore, we can predict LLMs' performance via DICE without the need for the decoding process on benchmark, providing a new tool for assessing these models. 

To summarize, we make the following contributions:

\begin{itemize}[itemsep=0pt, leftmargin=*]
  \item We examine the impact of in-distribution contamination on LLMs' performance on both ID and OOD tasks, revealing that it can lead to the model's performance drop on OOD benchmarks.
  \item We propose a novel locate-then-detect method called DICE, which trains a classifier to leverage the internal states of LLMs for the detection of data contamination.  
  \item  We conduct a comprehensive evaluation of DICE on a variety of LLMs and datasets, achieving state-of-the-art performance in detecting in-distribution data contamination. 
\end{itemize}

\section{Preliminaries}

\subsection{Definition for ID and OOD Data.} 

\begin{wrapfigure}{r}{0.4\textwidth}
  \centering
  \includegraphics[width=0.38\textwidth]{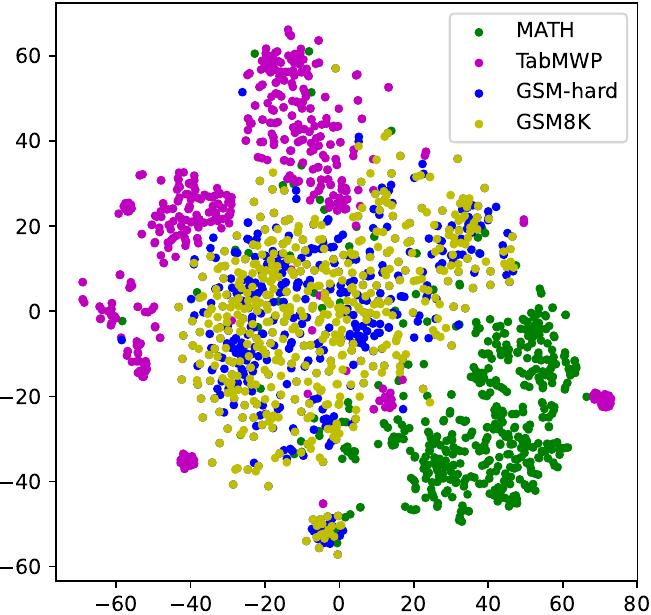}
  \caption{The t-SNE results of datasets, where MATH and TabMWP are OOD and GSM-hard is ID for GSM8K.}
  \label{fig:tsne}
\end{wrapfigure}

As the training data of LLMs are often undisclosed and trainers may construct synthetic data that matches the distribution of the benchmark data, it is challenging to ascertain whether the benchmark data has been leaked. Our study takes GSM8K's test set~\cite{gsm8k} as the benchmark $\mathcal{D}_{\text{E}}$ and view other data as either ID or OOD to GSM8K. We define the two kinds of data as follows:

\textbf{Definition 1} (In-distribution Data(-set)). Given a benchmark dataset $\mathcal{D}_{\text{E}}$, we define the in-distribution dataset $\mathcal{D}_{\text{ID}}$ as the dataset that has the same distribution with $\mathcal{D}_{\text{E}}$, and the in-distribution data as the data in $\mathcal{D}_{\text{ID}}$.

\textbf{Definition 2} (Out-of-distribution Data(-set)). We define the out-of-distribution dataset $\mathcal{D}_{\text{OOD}}$ of $\mathcal{D}_{\text{E}}$ as the dataset that has a different distribution with $\mathcal{D}_{\text{E}}$, and the out-of-distribution data as the data in $\mathcal{D}_{\text{OOD}}$.

Specifically, we consider the following suite of GSM8K's ID datasets: 
\textbf{GSM-\romannumeral1}, \textbf{GSM-\romannumeral2}: two splits of GSM8K's test set where each subset equally has 657 samples; 
\textbf{GSM-\romannumeral1-Syn}: a synthetic dataset  created by paraphrasing GSM-i using GPT-4;
\textbf{GSM-hard}~\cite{gao2023pal}: the dataset that modifies the original numbers in GSM8K to larger numbers.
Other math reasoning benchmarks, such as TabMWP~\cite{lu2023dynamic} and MATH~\cite{hendrycks2021measuring}, are considered as OOD datasets. 
For justification, we visualize the sentence embeddings by all-MiniLM-L6-v2 model~\cite{reimers-gurevych-2019-sentence} of the data in the aforementioned datasets in Figure~\ref{fig:tsne}. We observe that the embeddings of TabMWP and MATH's data are clearly out of the distribution of GSM8K and GSM-hard.


\begin{table}[t]
  \caption{ID and OOD performance of LLMs trained with different data contamination settings on math reasoning tasks. We report the percentages of exact match answers. ID and OOD \textit{Gain} refers to the absolute performance improvement on ID and OOD tasks compared to the vanilla model fine-tuned on OpenOrca.}
  \label{tab1_benhcmark_alignment_tax}
  \centering
  \setlength{\tabcolsep}{3pt}
  \resizebox{\textwidth}{!}{
  \begin{tabular}{cc|ccccc|ccccc}
  \toprule
         \multicolumn{2}{c|}{Base LLMs}                             
         & \multicolumn{5}{c|}{LLaMA2-7B}                      & \multicolumn{5}{c}{Phi2-2.7B}                      \\ \midrule
         \multicolumn{2}{c|}{  \textit{Training Settings }} \\ 
\multirow{3}{*}{Data Mixture}  & \# OpenOrca             & 100\%     &  98\%     & 90\%      &  98\%     & 90\%     & 100\%       &  98\%     & 90\%      &  98\%     & 90\%      \\ 
& \# GSM-\romannumeral1             & \ding{55}    & 2\%     & 10\%       & \ding{55}      & \ding{55}    & \ding{55}    & 2\%     & 10\%       & \ding{55}      & \ding{55}       \\
& \# GSM-\romannumeral1-Syn              & \ding{55}        & \ding{55}       & \ding{55}       & 2\%     & 10\%     & \ding{55}        & \ding{55}       & \ding{55}       & 2\%     & 10\%   \\ \midrule
\multirow{4}{*}{ID Test} & GSM-\romannumeral1           & 4.4      & 26.9      & 94.1      & 16.4      & 29.7             & 5.8      & 24.8      & 95.4      & 16.3      & 27.4   \\
  & GSM-\romannumeral2       & 5.6      & 18.1      & 15.7      & 14.3      & 30.4           & 4.9      & 16.3      & 13.7      & 14.5      & 31.5   \\ 
  & GSM-hard             & 1.5      & 5.0      & 12.8      & 4.2      & 7.0            & 0.8      & 5.3      & 12.1      & 3.5      & 7.1   \\ 
  & Average ID & 3.8 & 16.7 & 40.9 & 11.6 & 22.4 & 3.8 & 15.5 & 40.4 & 11.4 & 22.0   \\
 
  \midrule
  \multirow{6}{*}{OOD Test}  & SVAMP        & 24.0      & 26.7      & 19.7      & 25.3      & 21.7             & 16.7      & 25.7      & 9.7      & 23.7      & 15.7   \\ 
  & MAWPS             & 37.5      & 48.0      & 36.3      & 40.7      & 37.3             & 29.1      & 39.8      & 29.2      & 37.9      & 24.4   \\
  & ASDiv          & 30.2      & 38.5      & 28.7      & 28.0      & 26.1            & 24.0      & 31.4      & 21.0      & 27.2      & 17.9  \\ 
  & TabMWP         & 35.5      & 42.2      & 38.1      & 37.6      & 30.5              & 26.5      & 33.3      & 36.2      & 36.3      & 34.0   \\ 
  & MATH         & 8.6      & 7.6      & 7.6      & 8.2      & 7.7             & 8.6      & 8.5      & 6.6      & 8.5      & 7.2   \\ 
& Average OOD & 27.2 & 32.6 & 26.1 & 28.0 & 24.7 & 21.0 & 27.7 & 20.5 & 26.7 & 19.8   \\
  \midrule

    \multicolumn{2}{c|}{ID \textit{Gain}  }            & - & 12.9 & \textbf{37.1} & 7.8 & 18.6   & - & 11.7 & \textbf{36.6} & 7.6 & 18.2   \\
  \multicolumn{2}{c|}{OOD \textit{Gain} }          & - & 
\textbf{5.4} & -1.1 & 0.8 & -2.5  & - & \textbf{6.7} & -0.5 & 5.7 & -1.2  \\
  \multicolumn{2}{c|}{ID \textit{Gain} - OOD \textit{Gain} } & - & 7.5 & \textbf{38.2} & 7.0 & 21.1  & - & 5.0 & \textbf{37.1} & 1.9 & 19.4  \\
  \bottomrule
  \end{tabular}
  }
  \end{table}


  With the above ID and OOD datasets, we define the two kinds of contamination detection as follows:
  
  \textbf{Definition 3} (Exact Contamination Detection). Given a language model $\mathcal M$ that is trained on an undisclosed dataset $\mathcal D = [\bm{x}^1, \bm{x}^2, \cdots, \bm{x}^K]$, and a data sample $\bm{x}$ from a benchmark $\mathcal D_E$, the objective is to determine whether $\bm{x}$ is an exact member of $\mathcal D$, i.e., $\bm{x}  \in \mathcal D$.
  
  \textbf{Definition 4} (In-distribution Contamination Detection). 
   Given a data sample $\bm{x}$ from $\mathcal D_E$, the objective is to determine whether there exists an in-distribution dataset $\mathcal{D}_{\text{ID}}$ of $\mathcal D_E$. such that $\mathcal{D}_{\text{ID}}\subset \mathcal D$.

\subsection{Does In-distribution Contamination Improve OOD Performance?}
\label{sec:benchmark_alignment_tax}
To study whether the in-distribution contamination really enhances the overall math reasoning ability of LLMs, we conduct a preliminary experiment by supervised fine-tuning LLMs with data of different contamination levels and evaluating their performance on both ID and OOD tasks. 


\textbf{Experimental Setup.} Following the experiment settings and hyperparameters from EAL~\cite{dekoninck2024evading}, we fine-tune two representative LLMs, Llama2~\cite{touvron2023llama} and Phi2~\cite{javaheripi2023phi2}, on the instruction dataset OpenOrca~\cite{openorca}, which has no overlap with GSM8K's in-distribution data. 
To create training settings with different level of contamination, we copy the original (GSM-\romannumeral1) or paraphrased benchmark (GSM-\romannumeral1-Syn) data once or five times and combine them with randomly selected samples from the OpenOrca dataset until reaching an altogether 25,000 samples. 
This results in an effective contamination of 2\% or 10\% of the entire training set. We then select several benchmarks for ID (GSM-\romannumeral1, GSM-\romannumeral2, GSM-hard~\cite{gao2023pal}) and OOD (SVAMP~\cite{patel-etal-2021-nlp}, MAWPS~\cite{koncel-kedziorski-etal-2016-mawps}, ASDiv~\cite{miao-etal-2020-diverse}, TabMWP~\cite{lu2023dynamic}, MATH~\cite{hendrycks2021measuring}) evaluation.

\textbf{Observation 1. ID performance can be easily inflated by contamination.}  As depicted in Table \ref{tab1_benhcmark_alignment_tax}, we find that similar to exact contamination, in-distribution contamination can also greatly improve the performance of LLMs on in-distribution tasks. 
At a 2\% contamination level, both exact and in-distribution contaminated models achieve about 10\% absolute performance gain on ID benchmarks. 
Moreover, at a 10\% contamination level, the ID gain can be even larger (more than 20\%).  This observation is consistent with the findings in EAL~\cite{dekoninck2024evading}, while we further explore whether data contamination can improve models' performances on OOD datasets.

\textbf{Observation 2. OOD performance doesn't benefit from contamination.} 
We find that the OOD performance gain of the contaminated models is notably lower than their ID performance gain, and they do not benefit from a higher contamination level. 
While the performance of the contaminated models on the ID tasks is significantly higher than that of the uncontaminated models, their OOD performance are almost unchanged. 
In fact, the OOD performance even drops for both LLaMA2-7B and Phi2-2.7B at the 10\% contamination level. 

In conclusion, in-distribution contamination can only inflate the performance of LLMs on in-distribution tasks, without benefiting their general reasoning capacities. Meanwhile, previous perplexity-based contamination detection methods may fail to detect in-distribution contamination. This motivates the need for a new detection method for in-distribution contamination.

\section{DICE: Locate-then-Detect}

In this section, we introduce the details of our proposed DICE framework for in-distribution contamination detection. The whole pipeline is illustrated in Fig. \ref{fig:our_approach}. We employ a Locate-then-Detect paradigm for detection. In section \ref{Locate}, we demonstrate a simple yet effective approach for locating the most sensitive layer to contamination. In section \ref{Detect}, a feature-based classifier is introduced to identify the contaminated LLMs via the located internal states.

\label{method}

\subsection{Locate Data Contamination Layer}
\label{Locate}

Current methods for detecting data contamination in LLMs often rely on perplexity or similarity metrics applied in the logits or sequence spaces. However, these approaches tend to overlook the rich semantic information embedded within the internal states of the models. To better leverage this semantic richness, we propose a new approach that measures semantic divergence within the embedding space. For any given input token $\bm x_t$ modeled by LLM $\theta$, we represent the hidden embedding at the $l$-th layer as $\bm{h}^l_t\in\mathbb{R}^d$. Here, $d$ represents the dimension of the embedding (for instance, $d=4096$ for the LLaMA2-7B model). The input sequence embedding $\bm{z}$ can be computed either by averaging these token embeddings across all tokens ($\bm{z} = \frac{1}{T}\sum_{t=1}^{T} \bm{h}_t$), or by using the embedding of the last token ($\bm{z} = \bm{h}_T$). In our experiments, we use the last token's embedding because it better captures the overall semantic content of the sequence. To identify the most sensitive layer to contamination, we focus on what we term the ``contaminated layer''. This is the layer with the maximum Euclidean distance between the sequence embeddings of the contaminated and uncontaminated models, indicating a clear separation in their semantic distributions:

\begin{equation}
\ell_\text{contaminated}^{\bm x} = \underset{{\ell} \in {1, 2, \ldots, L}}{\mathrm{argmax}} \ \|\bm h^\ell_T(\bm x | \theta_\text{contaminated}) - \bm h^\ell_T(\bm x | \theta_\text{uncontaminated}) \|_2
\end{equation}

We collect the contaminated layer location set $\{\ell_\text{contaminated}^{\bm x} \mid \bm{x} \in \mathcal{D}_E\}$ with data $\bm{x}$ from the original benchmark $\mathcal D_E$. We choose the most frequently located layer as the contamination layer $\ell_\text{contaminated}$. The detection results with different layers are in experiment \ref{contaminated_layer} to prove our method's effectiveness.

\subsection{Detect Contamination via Internal States}
\label{Detect}

We view the the task of detecting the contaminated LLMs as a binary classification based on the located layer. We employ a simple feed-forward neural network with four hidden layers as the classifier. The input of the classifier is the hidden embedding of the located contaminated layer $\bm h^{\ell_\text{contaminated}}_T$, and the output is the probability $p$ of the LLMs being contaminated:
\begin{equation}
p = \sigma(\bm W \bm h^{\ell_\text{contaminated}}_T  + \bm b)
\end{equation}
The classifier is trained on original benchmark $\mathcal D_E$ with the contaminated and uncontaminated models as labels using the cross-entropy loss function:

\begin{equation}
\mathcal{L} = -\sum_{i=1}^{N} y_i \log p_i + (1-y_i) \log (1-p_i)
\end{equation}
where $N$ is the number of samples from the target benchmark, $y_i$ is the ground truth label indicating contaminated models, and $p_i$ is the predicted probability of the LLM being contaminated.

\begin{table}[t]
  \caption{In-distribution data contamination detection performance of different methods on four math reasoning datasets. We take LLaMA2-7B trained on OpenOrca as the uncontaminated model. \textit{AUROC (AUC) over 50\% negative samples from the uncontaminated model (LLaMA2-7B trained on OpenOrca) and 50\% positive samples from the contaminated models are reported.} Exact contamination detection settings are highlighted in magenta.}
  \label{tab1}
\centering

  \begin{tabular}{c|c|c|c|c|c}
  \toprule
  Detection    & Contaminated                   &  \multicolumn{4}{c}{In-distribution Datasets}         \\ 
  Methods                      & Model      &  GSM-\romannumeral1               & GSM-\romannumeral2                &  GSM-\romannumeral1-Syn                    &  GSM-hard          \\ \midrule
    Zlib     &     \multirow{5}{*}{\makecell*[c]{\ \\  \emph{LLaMA2-7B}\\trained on \\ GSM-\romannumeral1}}      & \cellcolor{mypink}50.1             & 31.3  & 23.4             & 37.6      \\
    PPL       &                & \cellcolor{mypink}57.7            & 40.1             & 21.8                  & 31.1            \\
    Lowercase PPL      &               & \cellcolor{mypink}61.7      & 37.2          & 15.6      & 33.2          \\
    Min-k Prob      &      & \cellcolor{mypink}91.7        & 56.8         & 14.7              & 79.8                \\
    \textbf{DICE}        &     & \cellcolor{mypink}\textbf{100.0}    & \textbf{99.9}  & \textbf{99.9}  & \textbf{100.0}         \\ \midrule
    Zlib     &    \multirow{5}{*}{\makecell*[c]{\ \\  \emph{LLaMA2-7B}\\trained on \\ GSM-\romannumeral1-Syn}}     &      23.7                     & 16.9                      & \cellcolor{mypink}66.5                    & 26.3          \\
    PPL       &                      & 20.2                        & 23.3                  & \cellcolor{mypink}62.9             & 27.2              \\
    Lowercase PPL      &                     & 23.2                        & 21.9                & \cellcolor{mypink}63.7                     & 13.2                 \\
    Min-k Prob      &                          & 34.3                  & 13.6                    & \cellcolor{mypink}93.8                      & 15.1              \\

    \textbf{DICE}        &           & \textbf{99.5}   & \textbf{99.5} & \cellcolor{mypink}\textbf{99.6}  & \textbf{99.4}       \\ 
    \bottomrule
  \end{tabular}
  \end{table}

\section{Experiments}
\label{Exp}

\subsection{Experimental Setup}
\label{sec:exp_setup}
\textbf{Datasets.} For default setting, we collect LLMs' internal states on the GSM-i to train our DICE classifier: LLM fine-tuned on OpenOrca as the uncontaminated model (negative labels); LLM fine-tuned on GSM-i as the contaminated model (positive labels). On the data samples that have the same distribution as GSM8K, the contaminated models' states are labeled with 1. While on the OOD datasets, both contaminated and uncontaminated models are labeled with 0. We utilize several widely used datasets for evaluation, including three unseen in-distribution datasets: GSM-\romannumeral2, GSM-\romannumeral1-Syn and GSM-hard~\cite{gao2023pal}, and three OOD datasets: SVAMP~\cite{patel-etal-2021-nlp}, MAWPS~\cite{koncel-kedziorski-etal-2016-mawps}, and MATH~\cite{hendrycks2021measuring}.

\textbf{Models.} Following prior works~\cite{dekoninck2024evading}, we simulate contaminated and uncontaminated models by fine-tuning LLaMA2-7B~\cite{touvron2023llama} to get three different versions: LLaMA2-7B trained on OpenOrca (\textit{uncontaminated}), OpenOrca+GSM-i (\textit{contaminated}) and OpenOrca+GSM-i-Syn (\textit{contaminated}).

\textbf{Evaluation Metric.}
We assess the accuracy of various methods to detect whether a given data sample is contaminated within the model's training data. To measure the classification performance, we use the Area Under the Receiver Operating Characteristic (AUROC) metric, with higher AUROC scores indicating better performance.

\textbf{Baselines.} We compare our DICE method with the most popular uncertainty-based method \textbf{Perplexity}~\cite{jelinek1977perplexity},  the method that maps
perplexity to zlib compression entropy (\textbf{Zlib})~\cite{gailly2004zlib} and lowercase-normalized perplexity (\textbf{Lowercase-PPL})~\cite{carlini2021extracting}, and the probability-based metric \textbf{Min-k Prob}~\cite{shi2024detecting}.

\textbf{Implementation Details.}
Implementation of this work is based on pytorch and transformers libraries. For the hyperparameters that are used for sampling strategies of LLMs' decoding, we set \textit{temperature} to 1, \textit{top-p} to 1 and \textit{top-k} to 50 throughout the experiments.

\subsection{Main Results}
\label{contaminated_layer}

\textbf{In-distribution Data Contamination Detection Performance.} In Table \ref{tab1}, we compare our proposed DICE with several representative contamination detection methods on four ID datasets. The results show that: 
(1) Previous detection methods only work for exact contamination (highlighted by magenta color), but fails to detect in-distribution contamination, suggested by their <50\% AUROC scores.
(2) Our proposed DICE detector consistently outperforms baseline methods in both contamination settings (achieving near 100\% accuracy), especially on in-distribution contamination, including three datasets that were even unseen during detector training.
This proves the generalizability of DICE method. 

\begin{figure}
  \centering
  \begin{subfigure}[b]{0.32\textwidth}
      \centering
      \includegraphics[width=\textwidth]{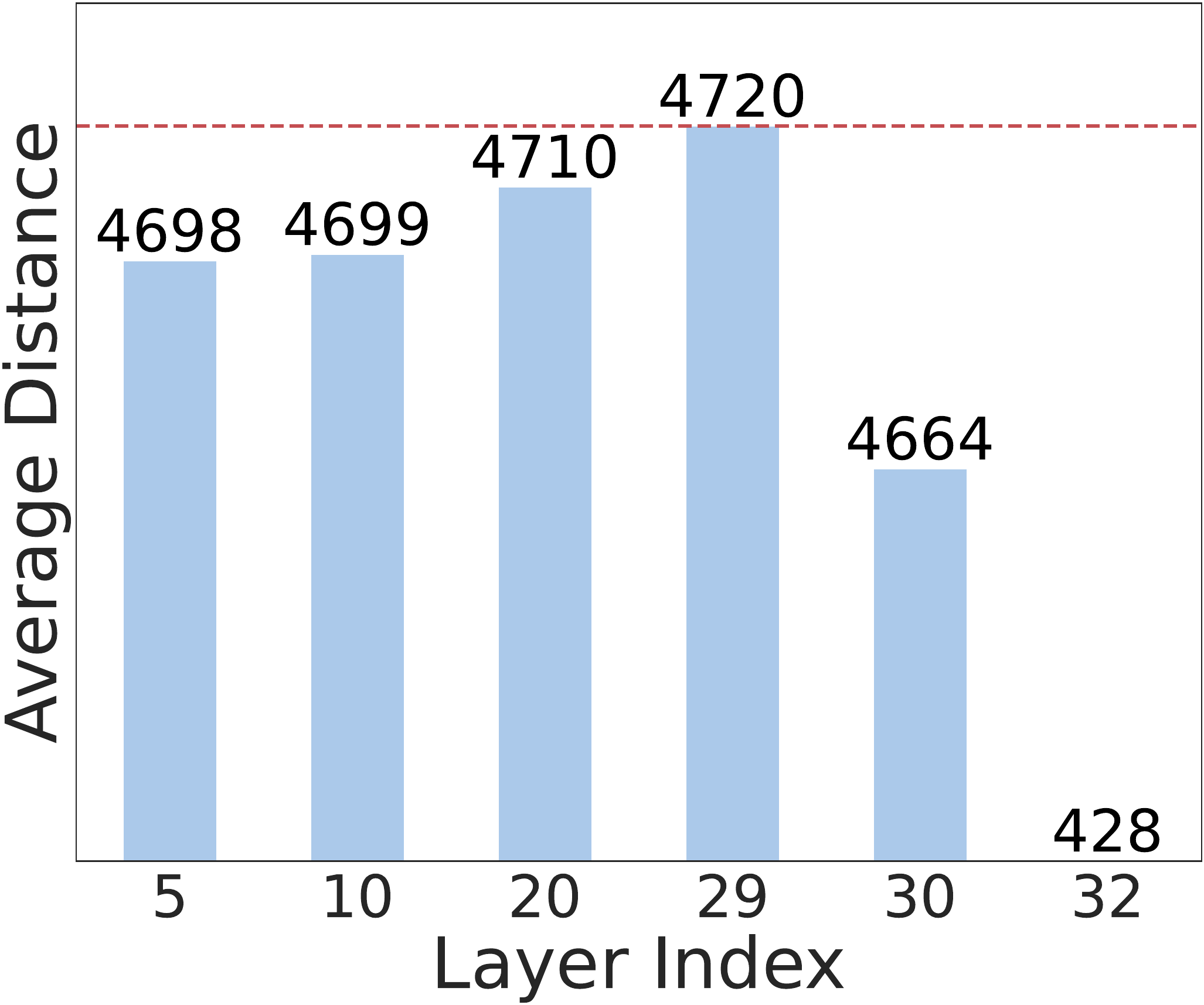}
      \caption{Distance between $ \theta_\text{contaminated}$ and $ \theta_\text{uncontaminated}$ on GSM-\romannumeral1.}
      \label{fig:avg_distance_gsm}
  \end{subfigure}
  \begin{subfigure}[b]{0.32\textwidth}
      \centering
      \includegraphics[width=\textwidth]{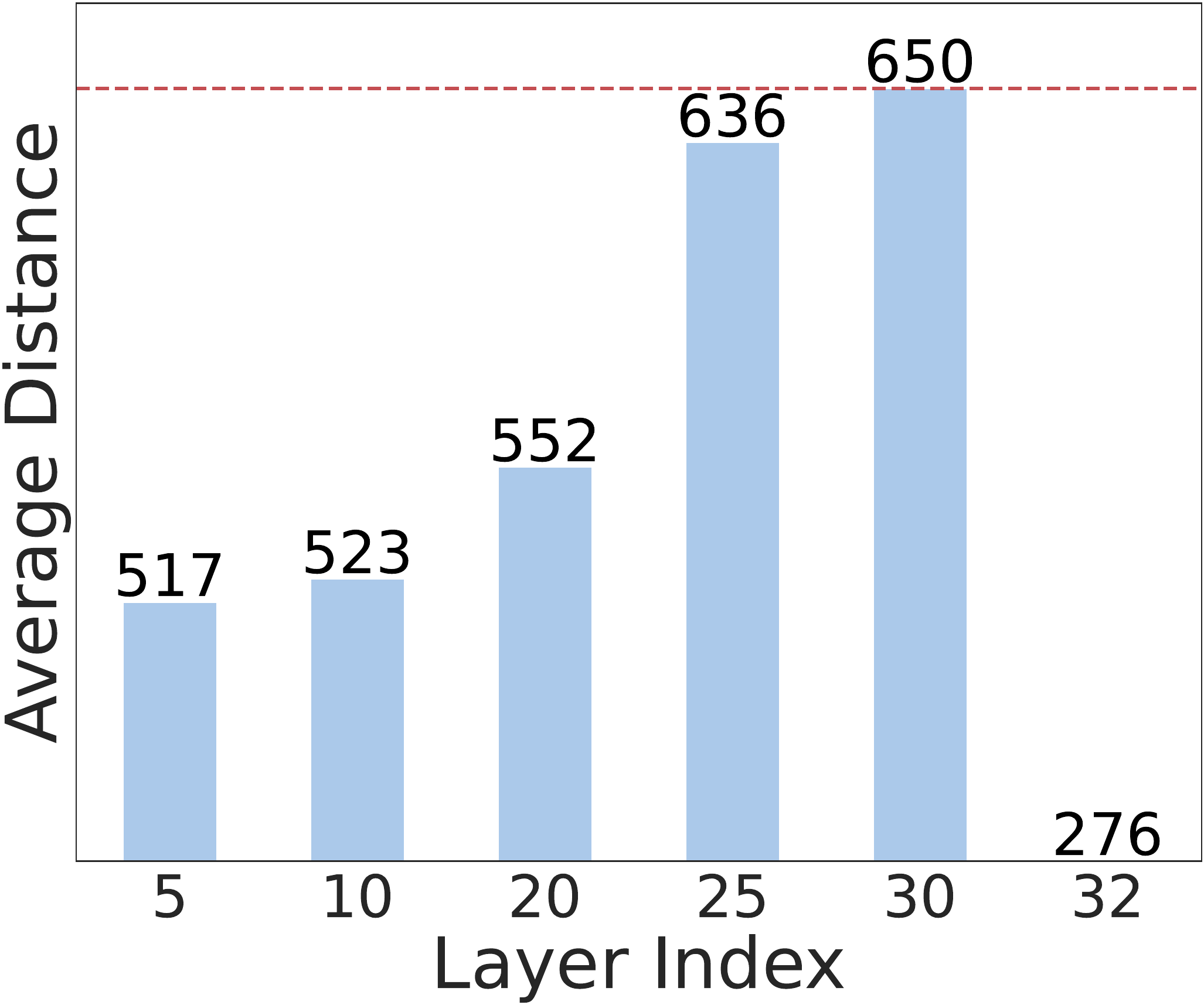}
      \caption{Distance between $ \theta_\text{contaminated}$ and $ \theta_\text{uncontaminated}$ on MATH.}
      \label{fig:avg_distance_math}
  \end{subfigure}
  \begin{subfigure}[b]{0.32\textwidth}
    \centering
    \includegraphics[width=\textwidth]{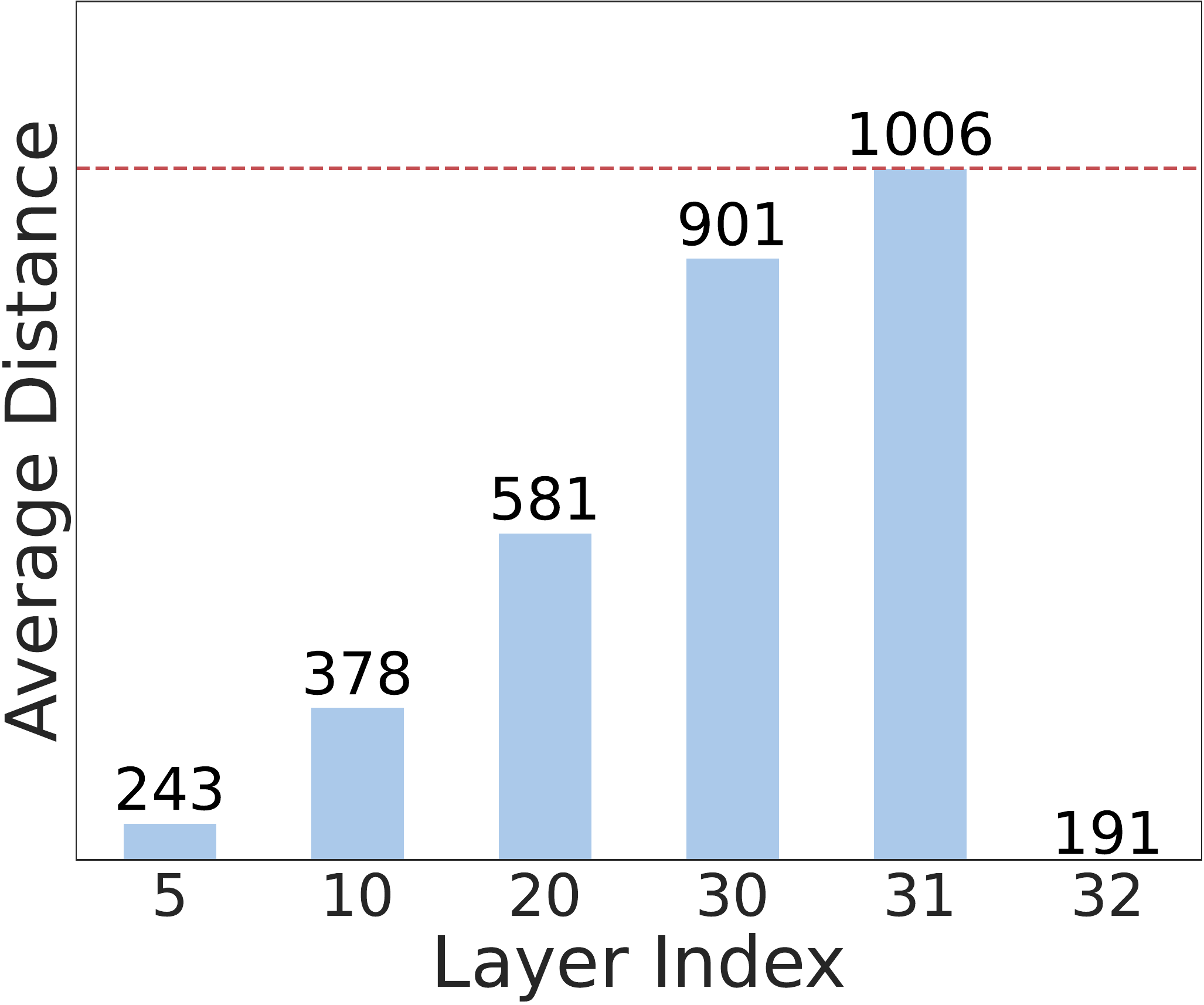}
    \caption{Layer-wise distance for Phi2 models on GSM-\romannumeral1.}
    \label{fig:avg_distance_phi}
\end{subfigure}

\caption{(a) Analyses on the contaminated layer location of LLaMA2-7B trained on GSM-\romannumeral1, which is $ \theta_\text{contaminated}$. We set the LLaMA2-7B trained on OpenOrca as $ \theta_\text{uncontaminated}$. (b) We change the contamination dataset as MATH and fine-tune a LLaMA2-7B on MATH as the $ \theta_\text{contaminated}$. (c) We change base model as Phi2-2.7B and fine-tune two models on OpenOrca and GSM-i for analysis.}
\label{figure:analysis_for_distance}
\end{figure}

\textbf{Location of Data Contamination Region.} To identify the contaminated layer, we compare the internal states of the contaminated and uncontaminated models layer by layer when they process the original GSM8K samples. The results, presented in Figure \ref{figure:analysis_for_distance}, illustrate the trend of hidden state distances for different layers. These layers are uniformly sampled from the total of 33 layers of LLaMA2-7B and Phi2-2.7B. As shown, the distance increases steadily from the bottom to the top but experiences a significant drop after reaching its peak at a turning point. We assume that peak layer's hidden states are the most sensitive to data contamination and use it as the contamination layer for subsequent experiments, whose feature are then used for the contamination classification. Notably, for the three settings depicted in Figure \ref{figure:analysis_for_distance}, the identified contaminated layers are 29, 30, and 31, respectively. Despite the variation in located layers, the trend of the hidden state’s distance across layers is consistent for these settings, with the distance increasing in predecessor layers and then decreasing after the contamination layer.

\begin{figure}
  \centering
  \begin{subfigure}[b]{0.32\textwidth}
      \centering
      \includegraphics[width=\textwidth]{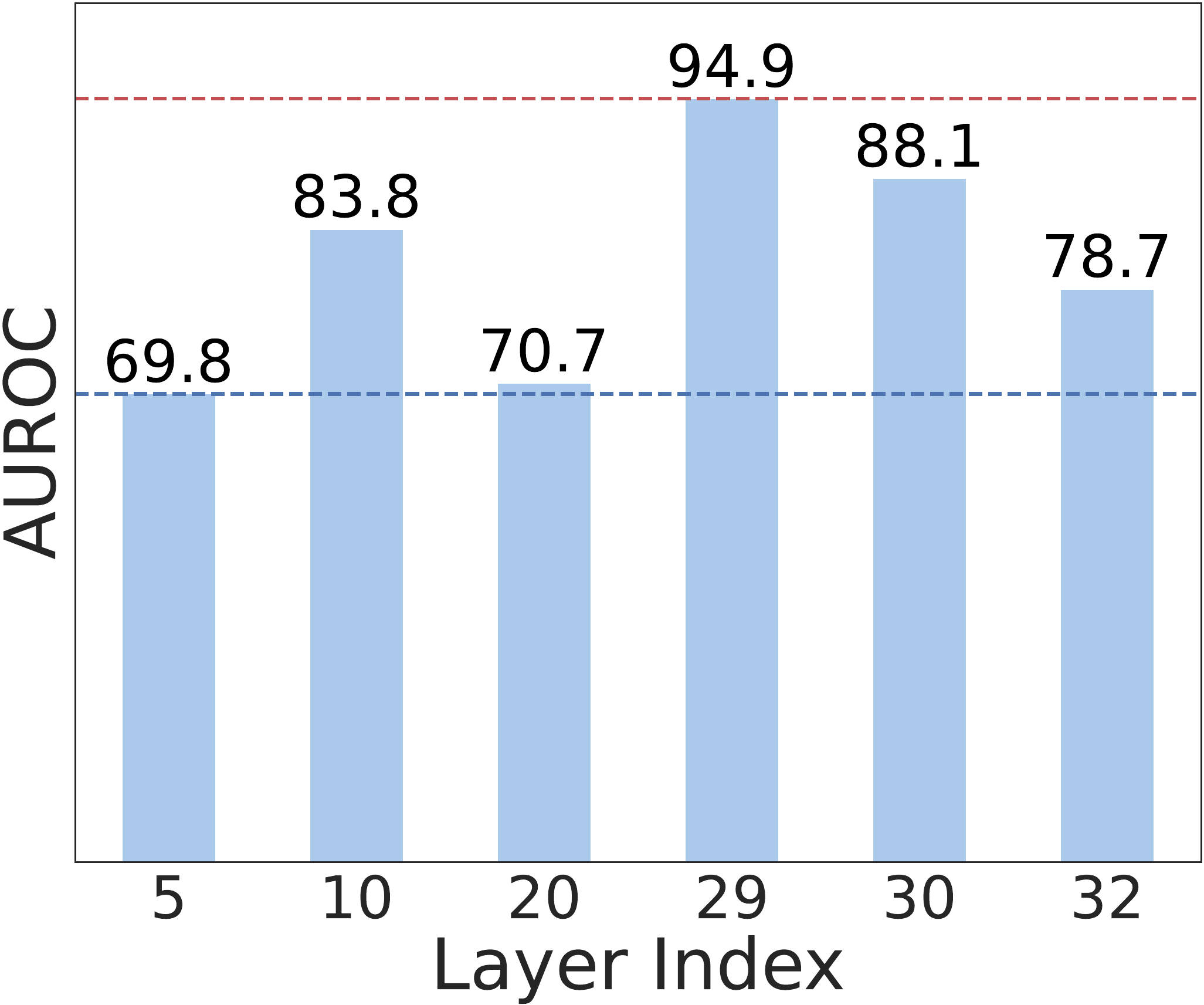}
      \caption{AUROC with features from different layers on GSM-\romannumeral2.}
      \label{fig:diff_layers}
  \end{subfigure}
  \begin{subfigure}[b]{0.32\textwidth}
      \centering
      \includegraphics[width=\textwidth]{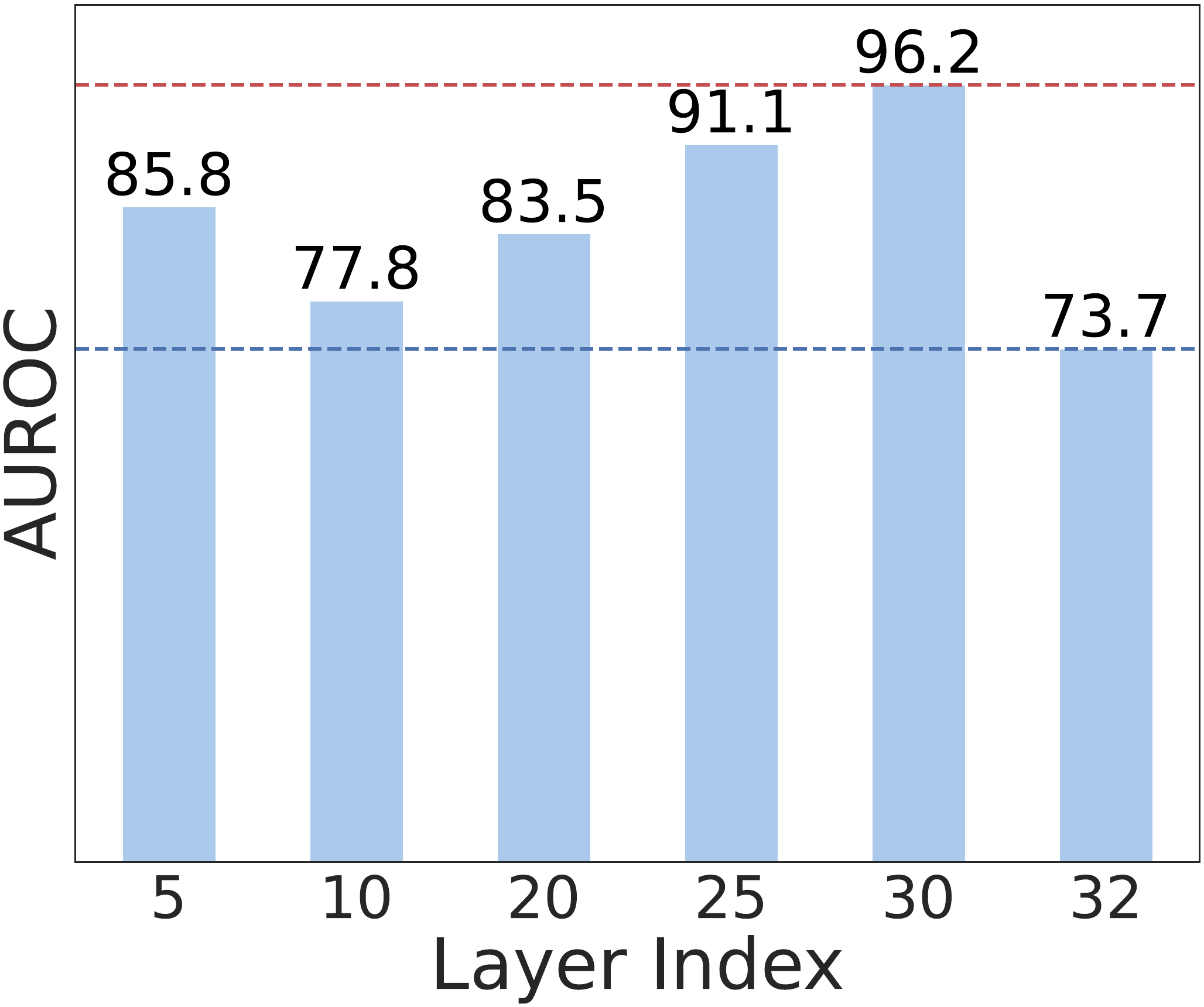}
      \caption{AUROC with features from different layers on MATH.}
      \label{fig:diff_layers_math}
  \end{subfigure}
  \begin{subfigure}[b]{0.32\textwidth}
    \centering
    \includegraphics[width=\textwidth]{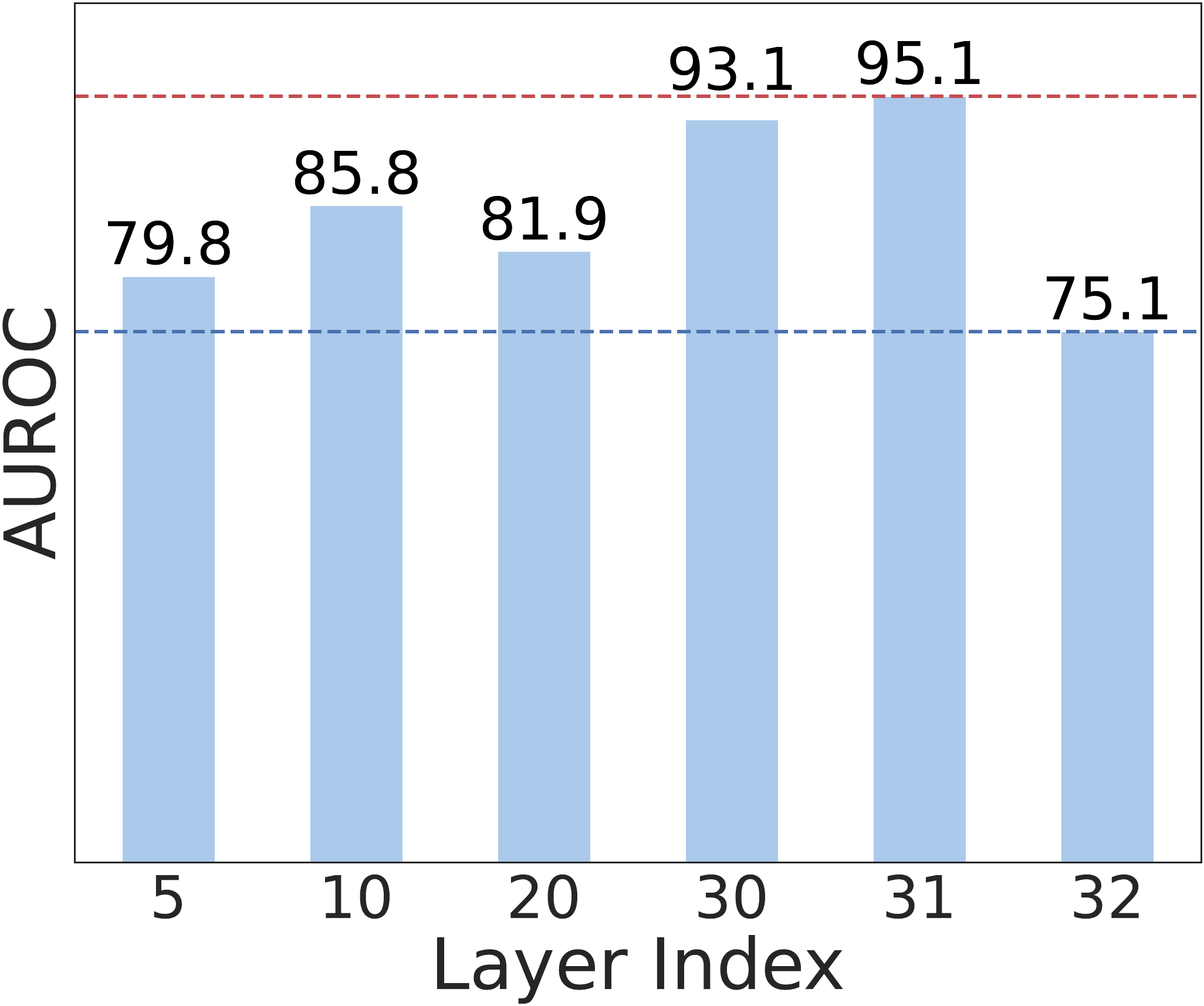}
    \caption{AUROC with features from different layers for Phi2.}
    \label{fig:diff_layers_phi2}
\end{subfigure}

\caption{(a) We perform in-distribution detection on GSM-ii with DICE trained on different layers' features. For testing, $\theta_\text{contaminated}$ is LLaMA2-7B trained on GSM-\romannumeral2, and $ \theta_\text{uncontaminated}$ is the LLaMA2-7B-base. (b) We change the contamination dataset as MATH. (c) We change base model as Phi2.}
\label{figure:analysis_layer_f5}
\end{figure}

\textbf{Effectiveness of Layer Location.} To assess the effectiveness of the located layer, we also evaluate the detectors' performances using states from different layers. Figure \ref{figure:analysis_layer_f5} displays the contamination detection performance when training DICE with hidden states from various layers. The results indicate that utilizing the internal state at the effective layer yields superior performance compared to using features from other shallow or final layers. Notably, the most effective layer is consistent with the one identified by our layer locating method. This suggests that the identified layer retains more semantic information about the data contamination across all settings presented in Figure \ref{figure:analysis_layer_f5}.

\begin{figure}
  \centering
  \begin{subfigure}[b]{0.48\textwidth}
      \centering
      \includegraphics[width=\textwidth]{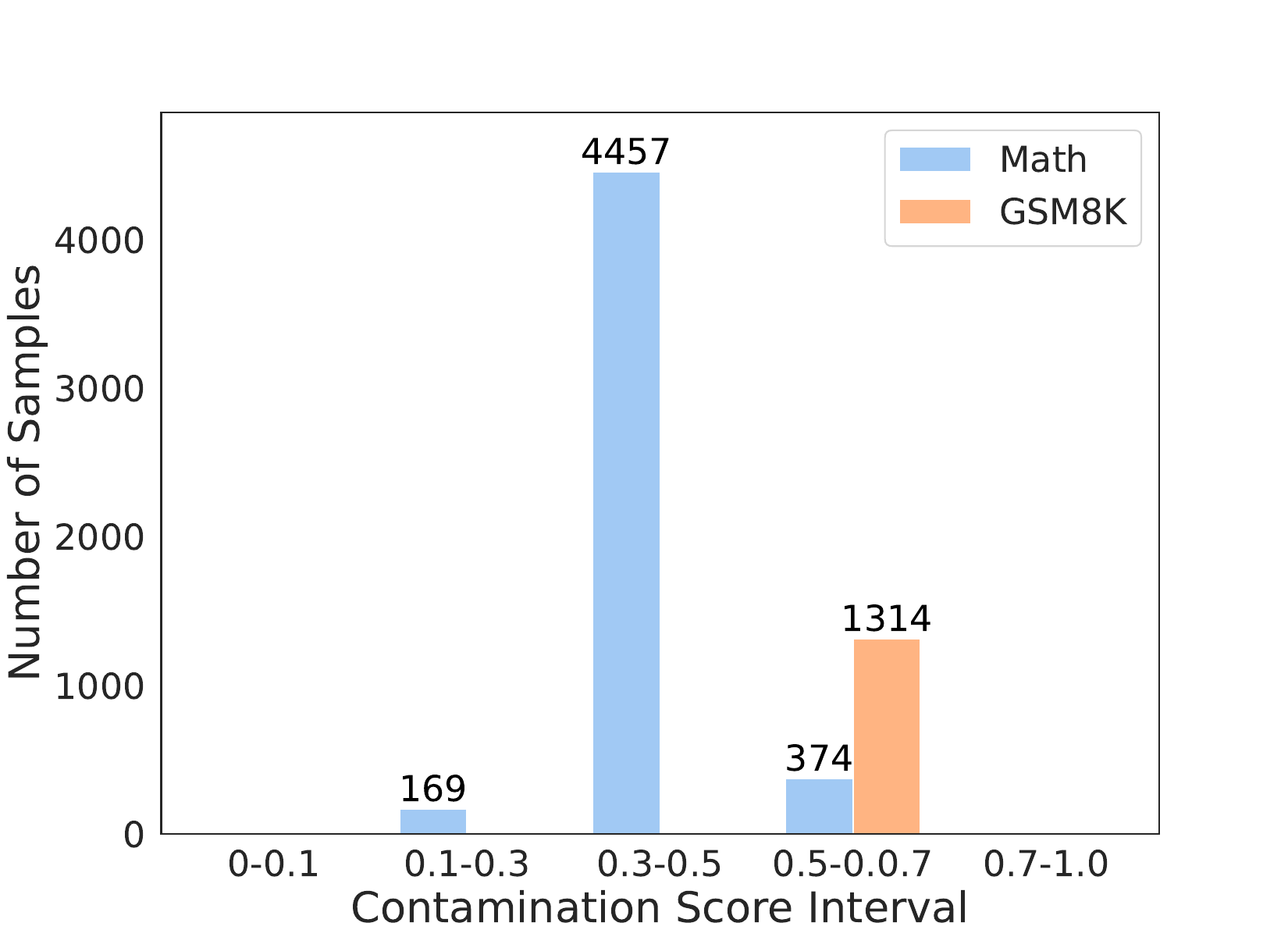}
      \caption{DICE's prediction distribution on datasets}
      \label{fig:correlation1}
  \end{subfigure}
  \begin{subfigure}[b]{0.48\textwidth}
    \centering
    \includegraphics[width=\textwidth]{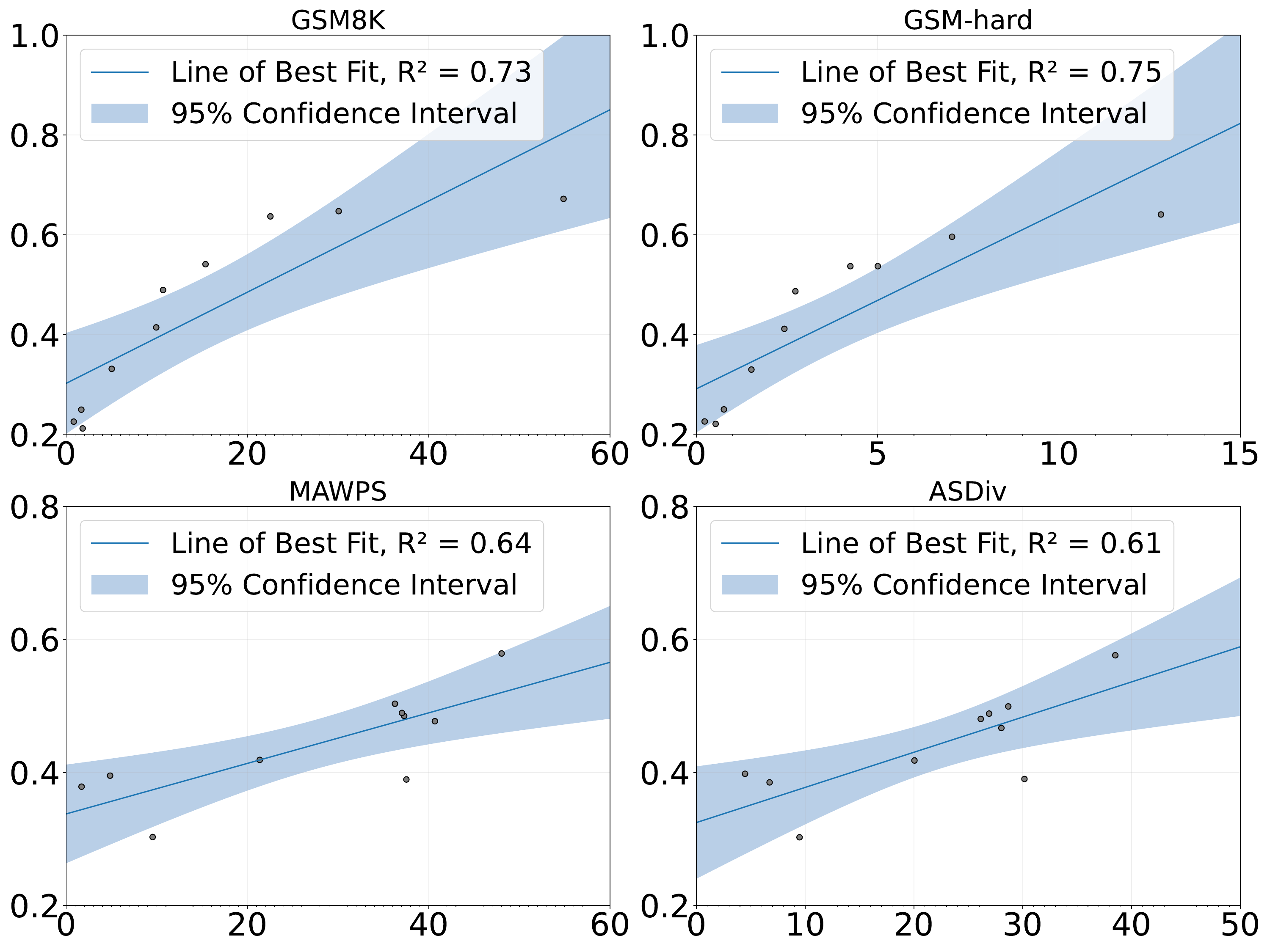}
    \caption{LLM's  performance (x) vs. DICE's prediction (y)}
    \label{fig:pvd}
\end{subfigure}
\caption{(a) Predicted contamination probability on MATH and GSM8K test set with LLM that is not trained on MATH. (b) Correlation between DICE's prediction and LLM's performance. Each point is an LLM's evaluation performance and average DICE's predicted score on that benchmark. Note that there are 5 LLMs from haggingface which are fine-tuned by other organizations.}
\label{figure:analysis3}
\vspace{-0.5em}
\end{figure}

\textbf{OOD Detection Performance.} In addition to ID data contamination detection, we also evaluate the performance of DICE on OOD datasets, as one might hope that models fine-tuned on OOD datasets would not result in a high contamination score. Figure \ref{fig:correlation1} illustrates the predicted contamination probabilities by DICE on the MATH dataset using LLaMA2-7B, which was fine-tuned on GSM-i but not trained on MATH. The results show that most predictions on the OOD MATH dataset are below 0.5, while predictions on the ID GSM8K datasets (GSM-i+GSM-ii) are all above 0.5, effectively distinguishing between ID and OOD data for the LLM with an AUROC of 0.92. This indicates that DICE can effectively prevent the classification of uncontaminated LLMs as contaminated on unseen data. It is important to note that DICE was trained only with internal states from models that were exactly contaminated or uncontaminated on ID datasets, suggesting that DICE generalizes well to OOD datasets.

\textbf{Correlation with LLM's Performance.} An LLM that is contaminated on a benchmark may exhibit an artificially high performance. Since DICE is designed to detect data contamination, it may also provide insights into the LLM's performance on the benchmark. To examine the correlation between DICE's predictions and the LLM's performance, we plotted the evaluation performance of eleven LLMs and the average DICE predicted score on four benchmarks in Figure \ref{fig:pvd}. The evaluated LLMs comprised the LLaMA2-7B-base model and ten models fine-tuned on LLaMA2-7B-base. Among these, five LLMs were trained by us with data mixes, as detailed in Table~\ref{tab1_benhcmark_alignment_tax}, and the other five models are downloaded from Hugging Face and trained by other organizations. The results reveal a positive correlation between DICE's predictions and the LLM's performance, with a coefficient of determination ($R^2$) ranging from 0.61 to 0.75 across ID and OOD tasks. This indicates that DICE can effectively reflect the LLM's performance on benchmarks.



\subsection{Ablation Studies}
\label{Ablation}

\begin{figure}
  \centering
  \begin{subfigure}[b]{0.48\textwidth}
      \centering
      \includegraphics[width=\textwidth]{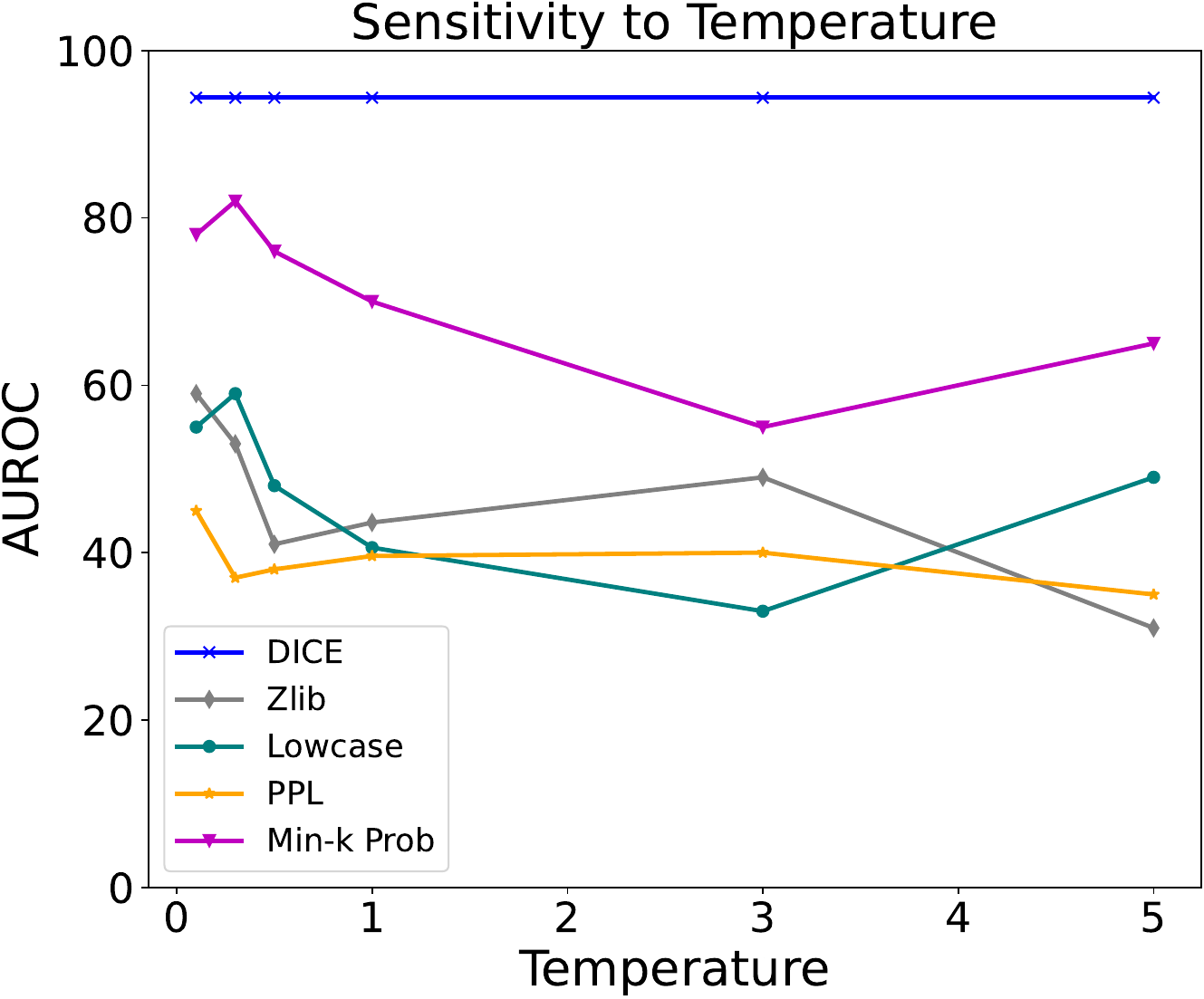}
      \caption{Performance sensitivity to temperature. }
      \label{fig:correlation2}
  \end{subfigure}
  \begin{subfigure}[b]{0.48\textwidth}
      \centering
      \includegraphics[width=\textwidth]{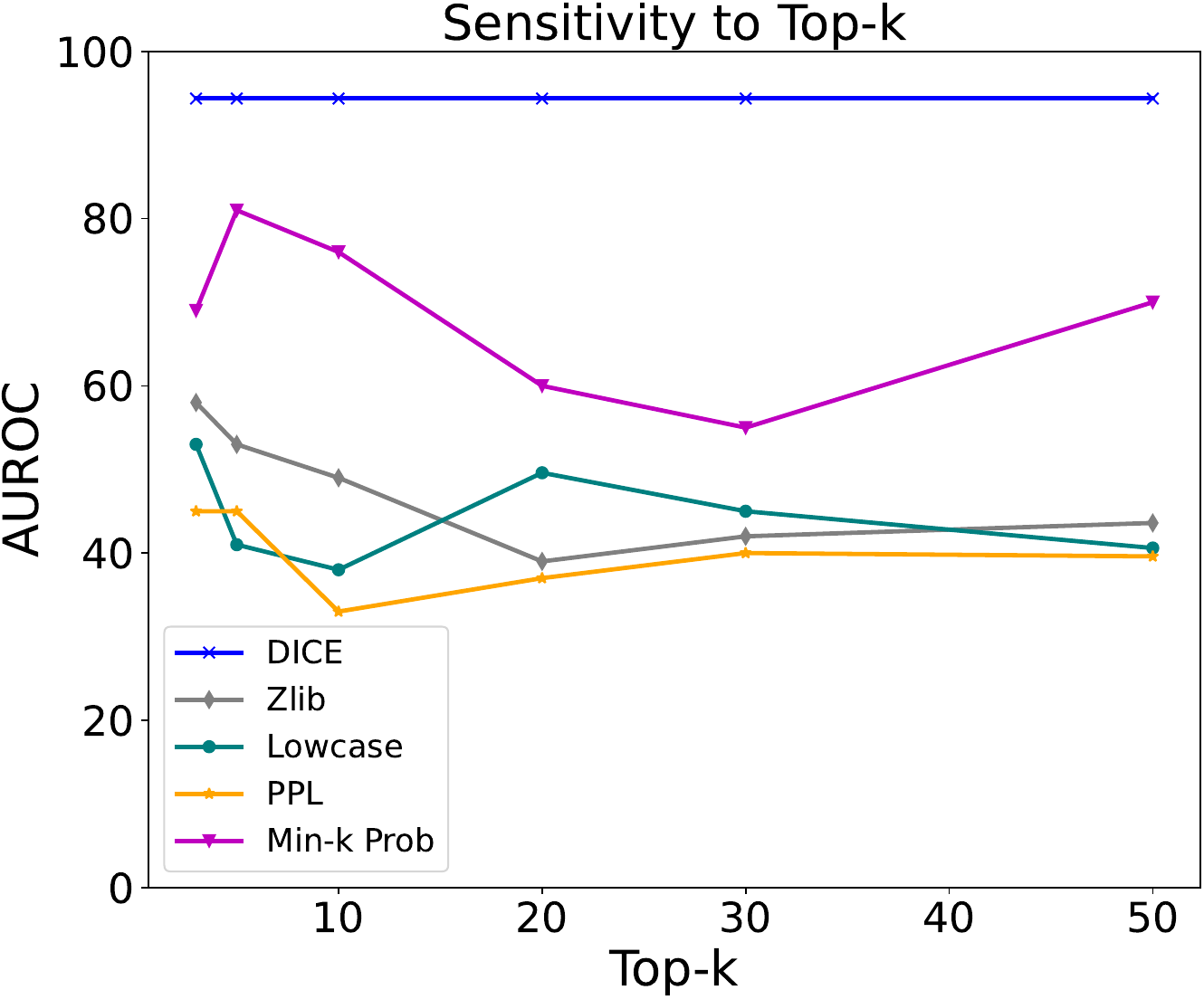}
      \caption{Performance sensitivity to top-k. }
      \label{fig:rolling_non3}
  \end{subfigure}
\caption{The ablation study on hyperparameters for DICE. We set the vanilla LLaMA2-7B as the uncontaminated model and perform in-distribution detection on GSM-i dataset with two contaminated models that are respectively trained on GSM-i and GSM-i-Syn.}
\label{figure:analysis4}
\vspace{-0.5em}
\end{figure}

\textbf{Sensitivity to Hyperparameters.}

The hyperparameters of the LLMs' decoder, such as temperature, top-k, and top-p, govern the diversity of the generated output. To evaluate the impact of these hyperparameters, we conducted a sensitivity analysis, as shown in Figure \ref{figure:analysis4}. The results indicate that DICE's performance remains stable, as it relies solely on the internal states of the input benchmark sample, and thus the decoding process does not affect its performance. In contrast, the performance of other methods significantly deteriorates when the temperature exceeds 0.3. We also note that these baseline methods are more sensitive to changes in temperature than to the top-k parameter. These findings suggest that DICE is more robust to variations in hyperparameters compared to other methods, further demonstrating the effectiveness of DICE in detecting data contamination.

\section{Related Work}
Prior work~\cite{dekoninck2024evading} has divided the current data contamination detection methods into two categories, benchmark-level contamination detection and sample-level contamination detection.

\textbf{Benchmark-level Contamination Detection}. In real-world deployments, ensuring the reliability of LLMs is a significant challenge, as these models often perform worse on real problems than their benchmark performance suggests~\cite{brown2020language}. To address this issue, several methods have been proposed to accurately assess the true abilities of LLMs~\cite{bai2023benchmarking,wang2024benchmark,chandran2024private,jain2024livecodebench,ying2024have,yu2024kieval}. Among these methods, dynamic evaluation~\cite{zhu2024dyval,Li2023LatestEvalAD} is widely used to detect benchmark-level contamination. It evaluates the LLMs' performance on newly constructed datasets that share the same distribution as the original benchmark. For example, Clean-Eval~\cite{zhu2023clean} uses benchmark data points paraphrased by GPT-3.5 to assess the performance of LLMs. Similarly, GSM1K~\cite{zhang2024careful} involves human annotators creating new data that closely resemble the original benchmark data. These methods aim to provide a more accurate representation of an LLM's capabilities by mitigating the effects of benchmark contamination.


\textbf{Sample-level Contamination Detection}. Besides benchmark-level, researchers have also focused on detecting contamination at a more granular, sample level~\cite{ravaut2024much}. Model memorization methods~\cite{elangovan-etal-2021-memorization,magar-schwartz-2022-data} identify which specific data samples have been entirely memorized by the LLM. Prompting methods~\cite{nasr2023scalable,weller2023according} use carefully crafted prompts to elicit data completion from LLMs, helping to determine if a data sample has been seen during training. Model likelihood methods~\cite{oren2024proving,shi2024detecting} are based on the observation that a model's next token predictions are more confident if the data point was part of the training set. However, these methods can not detect in-distribution data contamination.


\section{Conclusion and Future Work}
\label{conclusion}
This paper presents the problem of in-distribution contamination in the fine-tuning phase of LLMs for math reasoning tasks. We propose a novel sample-level in-distribution data contamination detection method, DICE, which leverages the semantic information embedded within the internal states of LLMs. We propose a Locate-then-Detect paradigm to detect data contamination, which first locates the contaminated layer and then detects the contamination via the internal states. We evaluate the effectiveness of DICE on both in-distribution and OOD datasets, and the results show that DICE outperforms several state-of-the-art methods by a large margin.  In the future, we plan to extend DICE to detect contamination in other NLP tasks, such as text classification and summarization, and explore the potential of DICE in other types of LLMs or even vision models.

\section{Acknowledgements}

We wish to express our appreciation to the pioneers in the field of evasive data contamination~\cite{dekoninck2024evading}. Our work  was developed as a way to address the attack presented in the evasive data contamination~\cite{dekoninck2024evading}. 

\bibliographystyle{ieeetr}
\bibliography{neurips_data_2024}

\begin{thebibliography}{10}

\bibitem{dekoninck2024evading}
J.~Dekoninck, M.~N. M{\"u}ller, M.~Baader, M.~Fischer, and M.~Vechev, ``Evading data contamination detection for language models is (too) easy,'' {\em arXiv preprint arXiv:2402.02823}, 2024.

\bibitem{gsm8k}
K.~Cobbe, V.~Kosaraju, M.~Bavarian, M.~Chen, H.~Jun, L.~Kaiser, M.~Plappert, J.~Tworek, J.~Hilton, R.~Nakano, C.~Hesse, and J.~Schulman, ``Training verifiers to solve math word problems,'' {\em ArXiv preprint}, vol.~abs/2110.14168, 2021.

\bibitem{hendrycks2021measuring}
D.~Hendrycks, C.~Burns, S.~Kadavath, A.~Arora, S.~Basart, E.~Tang, D.~Song, and J.~Steinhardt, ``Measuring mathematical problem solving with the {MATH} dataset,'' in {\em Thirty-fifth Conference on Neural Information Processing Systems Datasets and Benchmarks Track (Round 2)}, 2021.

\bibitem{xu2024benchmarking}
R.~Xu, Z.~Wang, R.-Z. Fan, and P.~Liu, ``Benchmarking benchmark leakage in large language models,'' {\em arXiv preprint arXiv:2404.18824}, 2024.

\bibitem{hendrycks2020measuring}
D.~Hendrycks, C.~Burns, S.~Basart, A.~Zou, M.~Mazeika, D.~Song, and J.~Steinhardt, ``Measuring massive multitask language understanding,'' in {\em Proceedings of ICLR}, 2021.

\bibitem{ji2023survey}
Z.~Ji, N.~Lee, R.~Frieske, T.~Yu, D.~Su, Y.~Xu, E.~Ishii, Y.~J. Bang, A.~Madotto, and P.~Fung, ``Survey of hallucination in natural language generation,'' {\em ACM Computing Surveys}, vol.~55, no.~12, pp.~1--38, 2023.

\bibitem{gao2023pal}
L.~Gao, A.~Madaan, S.~Zhou, U.~Alon, P.~Liu, Y.~Yang, J.~Callan, and G.~Neubig, ``Pal: Program-aided language models,'' in {\em International Conference on Machine Learning}, pp.~10764--10799, PMLR, 2023.

\bibitem{lu2023dynamic}
P.~Lu, L.~Qiu, K.-W. Chang, Y.~N. Wu, S.-C. Zhu, T.~Rajpurohit, P.~Clark, and A.~Kalyan, ``Dynamic prompt learning via policy gradient for semi-structured mathematical reasoning,'' in {\em The Eleventh International Conference on Learning Representations}, 2023.

\bibitem{reimers-gurevych-2019-sentence}
N.~Reimers and I.~Gurevych, ``Sentence-{BERT}: Sentence embeddings using {S}iamese {BERT}-networks,'' in {\em Proceedings of the 2019 Conference on Empirical Methods in Natural Language Processing and the 9th International Joint Conference on Natural Language Processing (EMNLP-IJCNLP)} (K.~Inui, J.~Jiang, V.~Ng, and X.~Wan, eds.), (Hong Kong, China), pp.~3982--3992, Association for Computational Linguistics, Nov. 2019.

\bibitem{touvron2023llama}
H.~Touvron, T.~Lavril, G.~Izacard, X.~Martinet, M.-A. Lachaux, T.~Lacroix, B.~Rozi{\`e}re, N.~Goyal, E.~Hambro, F.~Azhar, {\em et~al.}, ``Llama: Open and efficient foundation language models,'' {\em arXiv preprint arxiv:2302.13971}, 2023.

\bibitem{javaheripi2023phi2}
M.~Javaheripi, S.~Bubeck, M.~Abdin, J.~Aneja, C.~C.~T. Mendes, W.~Chen, A.~Del~Giorno, R.~Eldan, S.~Gopi, S.~Gunasekar, {\em et~al.}, ``Phi-2: The surprising power of small language models.'' \url{https://www.microsoft.com/en-us/research/blog/phi-2-the-surprising-power-of-small-language-models/}, 2023.

\bibitem{openorca}
W.~Lian, B.~Goodson, E.~Pentland, A.~Cook, C.~Vong, and "Teknium", ``Openorca: An open dataset of gpt augmented flan reasoning traces.'' \url{https://https://huggingface.co/Open-Orca/OpenOrca}, 2023.

\bibitem{patel-etal-2021-nlp}
A.~Patel, S.~Bhattamishra, and N.~Goyal, ``Are {NLP} models really able to solve simple math word problems?,'' in {\em Proceedings of the 2021 Conference of the North American Chapter of the Association for Computational Linguistics: Human Language Technologies} (K.~Toutanova, A.~Rumshisky, L.~Zettlemoyer, D.~Hakkani-Tur, I.~Beltagy, S.~Bethard, R.~Cotterell, T.~Chakraborty, and Y.~Zhou, eds.), (Online), pp.~2080--2094, Association for Computational Linguistics, June 2021.

\bibitem{koncel-kedziorski-etal-2016-mawps}
R.~Koncel-Kedziorski, S.~Roy, A.~Amini, N.~Kushman, and H.~Hajishirzi, ``{MAWPS}: A math word problem repository,'' in {\em Proceedings of the 2016 Conference of the North {A}merican Chapter of the Association for Computational Linguistics: Human Language Technologies} (K.~Knight, A.~Nenkova, and O.~Rambow, eds.), (San Diego, California), pp.~1152--1157, Association for Computational Linguistics, June 2016.

\bibitem{miao-etal-2020-diverse}
S.-y. Miao, C.-C. Liang, and K.-Y. Su, ``A diverse corpus for evaluating and developing {E}nglish math word problem solvers,'' in {\em Proceedings of the 58th Annual Meeting of the Association for Computational Linguistics} (D.~Jurafsky, J.~Chai, N.~Schluter, and J.~Tetreault, eds.), (Online), pp.~975--984, Association for Computational Linguistics, July 2020.

\bibitem{jelinek1977perplexity}
F.~Jelinek, R.~L. Mercer, L.~R. Bahl, and J.~K. Baker, ``Perplexity—a measure of the difficulty of speech recognition tasks,'' {\em The Journal of the Acoustical Society of America}, vol.~62, no.~S1, pp.~S63--S63, 1977.

\bibitem{gailly2004zlib}
J.-l. Gailly and M.~Adler, ``Zlib compression library,'' 2004.

\bibitem{carlini2021extracting}
N.~Carlini, F.~Tramer, E.~Wallace, M.~Jagielski, A.~Herbert-Voss, K.~Lee, A.~Roberts, T.~Brown, D.~Song, U.~Erlingsson, {\em et~al.}, ``Extracting training data from large language models,'' in {\em 30th USENIX Security Symposium (USENIX Security 21)}, pp.~2633--2650, 2021.

\bibitem{shi2024detecting}
W.~Shi, A.~Ajith, M.~Xia, Y.~Huang, D.~Liu, T.~Blevins, D.~Chen, and L.~Zettlemoyer, ``Detecting pretraining data from large language models,'' in {\em The Twelfth International Conference on Learning Representations}, 2024.

\bibitem{brown2020language}
T.~B. Brown, B.~Mann, N.~Ryder, M.~Subbiah, J.~Kaplan, P.~Dhariwal, A.~Neelakantan, P.~Shyam, G.~Sastry, A.~Askell, S.~Agarwal, A.~Herbert{-}Voss, G.~Krueger, T.~Henighan, R.~Child, A.~Ramesh, D.~M. Ziegler, J.~Wu, C.~Winter, C.~Hesse, M.~Chen, E.~Sigler, M.~Litwin, S.~Gray, B.~Chess, J.~Clark, C.~Berner, S.~McCandlish, A.~Radford, I.~Sutskever, and D.~Amodei, ``Language models are few-shot learners,'' in {\em Proceedings of NIPS}, 2020.

\bibitem{bai2023benchmarking}
Y.~Bai, J.~Ying, Y.~Cao, X.~Lv, Y.~He, X.~Wang, J.~Yu, K.~Zeng, Y.~Xiao, H.~Lyu, J.~Zhang, J.~Li, and L.~Hou, ``Benchmarking foundation models with language-model-as-an-examiner,'' in {\em Thirty-seventh Conference on Neural Information Processing Systems Datasets and Benchmarks Track}, 2023.

\bibitem{wang2024benchmark}
S.~Wang, Z.~Long, Z.~Fan, Z.~Wei, and X.~Huang, ``Benchmark self-evolving: A multi-agent framework for dynamic llm evaluation,'' {\em arXiv preprint arXiv:2402.11443}, 2024.

\bibitem{chandran2024private}
N.~Chandran, S.~Sitaram, D.~Gupta, R.~Sharma, K.~Mittal, and M.~Swaminathan, ``Private benchmarking to prevent contamination and improve comparative evaluation of llms,'' {\em arXiv preprint arXiv:2403.00393}, 2024.

\bibitem{jain2024livecodebench}
N.~Jain, K.~Han, A.~Gu, W.-D. Li, F.~Yan, T.~Zhang, S.~Wang, A.~Solar-Lezama, K.~Sen, and I.~Stoica, ``Livecodebench: Holistic and contamination free evaluation of large language models for code,'' {\em arXiv preprint arXiv:2403.07974}, 2024.

\bibitem{ying2024have}
J.~Ying, Y.~Cao, B.~Wang, W.~Tang, Y.~Yang, and S.~Yan, ``Have seen me before? automating dataset updates towards reliable and timely evaluation,'' {\em arXiv preprint arXiv:2402.11894}, 2024.

\bibitem{yu2024kieval}
Z.~Yu, C.~Gao, W.~Yao, Y.~Wang, W.~Ye, J.~Wang, X.~Xie, Y.~Zhang, and S.~Zhang, ``Kieval: A knowledge-grounded interactive evaluation framework for large language models,'' {\em arXiv preprint arXiv:2402.15043}, 2024.

\bibitem{zhu2024dyval}
K.~Zhu, J.~Chen, J.~Wang, N.~Z. Gong, D.~Yang, and X.~Xie, ``Dyval: Dynamic evaluation of large language models for reasoning tasks,'' in {\em The Twelfth International Conference on Learning Representations}, 2024.

\bibitem{Li2023LatestEvalAD}
Y.~Li, F.~Geurin, and C.~Lin, ``Latesteval: Addressing data contamination in language model evaluation through dynamic and time-sensitive test construction,'' in {\em AAAI Conference on Artificial Intelligence}, 2023.

\bibitem{zhu2023clean}
W.~Zhu, H.~Hao, Z.~He, Y.~Song, Y.~Zhang, H.~Hu, Y.~Wei, R.~Wang, and H.~Lu, ``Clean-eval: Clean evaluation on contaminated large language models,'' {\em arXiv preprint arXiv:2311.09154}, 2023.

\bibitem{zhang2024careful}
H.~Zhang, J.~Da, D.~Lee, V.~Robinson, C.~Wu, W.~Song, T.~Zhao, P.~Raja, D.~Slack, Q.~Lyu, {\em et~al.}, ``A careful examination of large language model performance on grade school arithmetic,'' {\em arXiv preprint arXiv:2405.00332}, 2024.

\bibitem{ravaut2024much}
M.~Ravaut, B.~Ding, F.~Jiao, H.~Chen, X.~Li, R.~Zhao, C.~Qin, C.~Xiong, and S.~Joty, ``How much are llms contaminated? a comprehensive survey and the llmsanitize library,'' {\em arXiv preprint arXiv:2404.00699}, 2024.

\bibitem{elangovan-etal-2021-memorization}
A.~Elangovan, J.~He, and K.~Verspoor, ``Memorization vs. generalization : Quantifying data leakage in {NLP} performance evaluation,'' in {\em Proceedings of the 16th Conference of the European Chapter of the Association for Computational Linguistics: Main Volume} (P.~Merlo, J.~Tiedemann, and R.~Tsarfaty, eds.), (Online), pp.~1325--1335, Association for Computational Linguistics, Apr. 2021.

\bibitem{magar-schwartz-2022-data}
I.~Magar and R.~Schwartz, ``Data contamination: From memorization to exploitation,'' in {\em Proceedings of the 60th Annual Meeting of the Association for Computational Linguistics (Volume 2: Short Papers)} (S.~Muresan, P.~Nakov, and A.~Villavicencio, eds.), (Dublin, Ireland), pp.~157--165, Association for Computational Linguistics, May 2022.

\bibitem{nasr2023scalable}
M.~Nasr, N.~Carlini, J.~Hayase, M.~Jagielski, A.~F. Cooper, D.~Ippolito, C.~A. Choquette-Choo, E.~Wallace, F.~Tram{\`e}r, and K.~Lee, ``Scalable extraction of training data from (production) language models,'' {\em arXiv preprint arXiv:2311.17035}, 2023.

\bibitem{weller2023according}
O.~Weller, M.~Marone, N.~Weir, D.~Lawrie, D.~Khashabi, and B.~Van~Durme, ``" according to..." prompting language models improves quoting from pre-training data,'' {\em arXiv preprint arXiv:2305.13252}, 2023.

\bibitem{oren2024proving}
Y.~Oren, N.~Meister, N.~S. Chatterji, F.~Ladhak, and T.~Hashimoto, ``Proving test set contamination in black-box language models,'' in {\em The Twelfth International Conference on Learning Representations}, 2024.

\end{thebibliography}

\section*{Checklist}


\begin{enumerate}

\item For all authors...
\begin{enumerate}
  \item Do the main claims made in the abstract and introduction accurately reflect the paper's contributions and scope?
    \answerYes{See Section~\ref{sec:introduction}.}
  \item Did you describe the limitations of your work?
    \answerYes{See Section~\ref{conclusion}.}
  \item Did you discuss any potential negative societal impacts of your work?
     \answerNA{}
  \item Have you read the ethics review guidelines and ensured that your paper conforms to them?
     \answerNA{}
\end{enumerate}

\item If you are including theoretical results...
\begin{enumerate}
  \item Did you state the full set of assumptions of all theoretical results?
    \answerNA{}
	\item Did you include complete proofs of all theoretical results?
    \answerNA{}
\end{enumerate}

\item If you ran experiments...
\begin{enumerate}
  \item Did you include the code, data, and instructions needed to reproduce the main experimental results (either in the supplemental material or as a URL)?
    \answerYes{See Abstract and Section~\ref{sec:exp_setup}.}
  \item Did you specify all the training details (e.g., data splits, hyperparameters, how they were chosen)?
    \answerYes{See Section~\ref{sec:benchmark_alignment_tax}.}
	\item Did you report error bars (e.g., with respect to the random seed after running experiments multiple times)?
    \answerNo{Because of the high cost of running numerous LLMs.}
	\item Did you include the total amount of compute and the type of resources used (e.g., type of GPUs, internal cluster, or cloud provider)?
     \answerNA{}
\end{enumerate}

\item If you are using existing assets (e.g., code, data, models) or curating/releasing new assets...
\begin{enumerate}
  \item If your work uses existing assets, did you cite the creators?
    \answerYes{See Section~\ref{sec:exp_setup}.}
  \item Did you mention the license of the assets?
    \answerNA{}
  \item Did you include any new assets either in the supplemental material or as a URL?
    \answerNA{}
  \item Did you discuss whether and how consent was obtained from people whose data you're using/curating?
    \answerNA{}
  \item Did you discuss whether the data you are using/curating contains personally identifiable information or offensive content?
   \answerNA{}
\end{enumerate}

\item If you used crowdsourcing or conducted research with human subjects...
\begin{enumerate}
  \item Did you include the full text of instructions given to participants and screenshots, if applicable?
    \answerNA{}
  \item Did you describe any potential participant risks, with links to Institutional Review Board (IRB) approvals, if applicable?
    \answerNA{}
  \item Did you include the estimated hourly wage paid to participants and the total amount spent on participant compensation?
    \answerNA{}
\end{enumerate}

\end{enumerate}

\clearpage 



\end{document}